# A fuzzy logic-based stabilization system for a flying robot, with an embedded energy harvester and a visual decision-making system

Abdullatif BABA [1a], Basel ALOTHMAN[b]

[a]*Kuwait College of Science and Technology, Computer Science and Engineering Department, Kuwait*
[a]*University of Turkish Aeronautical Association, Computer Engineering Department, Ankara, Turkey*
[b]*Kuwait College of Science and Technology, Computer Science and Engineering Department, Kuwait*

## ABSTRACT

"Smart cities" is the trendy rubric of modern urban projects that require new innovative ideas to attain perfection in many fields to improve our lives. In this context, a new innovative application will be presented here to investigate and continuously make the required maintenance of public roads by creating a flying robot for painting the partially erased parts of sidewalks' edges that are usually plated in two different colors; primarily black and white as we suppose here. The first contribution of this paper is developing a fuzzy-logic-based stabilization system for an octocopter serving as a liquids transporter that could be equipped with a robot arm. The second contribution consists of designing an embedded energy harvester for the flying robot to promote the management of available power sources. Finally, as suggested in this project, we present a complement heuristic study clarifying some main concepts that rely on a computer vision-based decision-making system.

## 1. Introduction

This paper introduces a fuzzy-logic-based design to stabilize an octocopter serving as a multi-purpose spraying machine that could be utilized in different areas like firefighting, spreading pesticides on infected plants, or performing façades coloring works. The robot structure suggested here contains a 3DoF robot arm, which may lead due to its arbitrary movement in all directions to get an asymmetric flying architecture. On the other hand, the robot is supposed to carry a reservoir of liquid material. Hence, the robot's motion will vibrate both; the robot elements and the transported liquid which in turn induces a considerable amount of additional vibrations that affect the total balance of the robot. In other words, we describe a closed-loop vibration scenario between two connected entities (the robot and the liquid). Avoiding this problem may require offering a few classical ideas, including the use of mechanical dampers, splitting the reservoir into circular cells connected by narrow orifices, or utilizing special containers that keep constant and continuous pressure on the transposed liquid even while it is spread out. In any case, one of these solutions should be adopted as an essential procedure. But, to attain a perfect balance of our design, we suggest and simulate a new technique that relies on a fuzzy logic algorithm. From a technical point of view and compared to a quadcopter, an octocopter could be regarded as a power-consuming device but simultaneously much more stable, Aspragkathos et al. (2022). This remark will lead us to develop the power management system for our device. Therefore, we present a new design of an embedded electromagnetic energy harvester when we discuss the technical specifications of this robot. Then, as a supplement heuristic study, we present a computer vision-based decision-making system that profits from the fundamental idea of the Hopfield network to detect and paint the outer side of sidewalks as an advanced automated strategy to be always accomplished by using flying robots for keeping a good appearance of public roads.

## 2. Related works

Flying robots will increasingly impact drawing intelligent features for our life. In this context, diverse designs of fixed-wing drones or multi-rotor UAVs were recently proposed to perform a wide range of smart applications in different topics, Radoglou-Grammatikis et al. (2020).
A quadcopter-based flying robot was suggested by Baba (2022) to investigate the state of electric power transmission systems and to execute any required maintenance procedure if a defect was identified. In the last few years, agricultural automation projects have vastly relied on this technology to increase the productivity of economic crops and reduce their operation costs, Albiero et al. (2022) and Guzman et al. (2021). A smart transportation system was suggested by Patchou et al. (2021) to encourage machine-human interaction, especially for dealing with pandemic circumstances like Covid-19. On the other hand, a mechanical design for damping a firefighting flying robot's oscillations has been suggested by Yamaguchi et al. (2019). While in another study Petritoli and Leccese (2020), fuzzy logic was used to simplify the model for longitudinal stability of a fixed-wing UAV to approximate human behavior and avoid human frailties. Another fuzzy logic-based controller was used by Petritoli and Leccese (2020) as a full command system for a hexadrone that relies on GPS signals with a reading error of around 40 cm as a linear distance. A hybrid controller could be noticed in Santoso et al. (2021) that takes into account PD and Fuzzy logic to track the trajectory of a quad-

a.baba@kcst.edu.kw; ababa@thk.edu.tr (A.B. );
b.alothman@kcst.edu.kw (B. ALOTHMAN)
ORCID(s): 0000-0001-5165-4205 (A.B. ); 0000-0002-2752-4811 (B. ALOTHMAN)

copter. Once again, Fuzzy logic will be also exploited by Kamil and Moghrabiah (2021) to navigate a mobile robot in a dynamic and non-deterministic environment, the objective of this study is to build a multilayer decision model. Recently, the stability of a humanoid robot was achieved by using a Fuzzy logic-based regulator to overcome two main problems; the uncertain changes in the center of mass height for zero moment point control and the variable stiffness of impedance control, Dong et al. (2022). Moreover, a fast and soft landing system for a quadcopter was presented by Kumar (2020), they used a Fuzzy logic-based technique. A swarm robotics study Barawkar and Kumar (2021) that concentrates on multi-drone cooperative transport, has also utilized Fuzzy logic to create a force-torque feedback controller that compensates for the offset effect of the known center of gravity. The rest of this article will be presented as follows; the next section describes all the relevant details about the design of our flying robot, including its main technical specifications, the electromagnetic-based embedded energy harvester, and the fuzzy logic-based stabilization system. Then, the computer vision-based decision-making system will be presented. Finally, we conclude with an overview of the main features of our design.

## 3. A full description of the design

The flying robot presented in this study, as shown in figure (1), is an octocopter equipped with different types of sensors like an integrated IMU&GPS sensor and an RGB camera. As it contains eight rotors, an octocopter could be considered a power-consuming device which may lead to incorporating two rechargeable Lipo batteries (10000 mAh) into the design. Contrariwise, due to its octagonal or "quasi-circular" arrangement of the eight rotors, an octocopter is always regarded as a much more stable device compared with a quadcopter. In such a case, developing our robot's power management system becomes a must. In this context, we suggest using an electromagnetic energy harvester as presented in the next paragraph.

### 3.1. The robot's technical specifications

All the required elements for building a flying robot for carrying and transporting 10 liters of liquid are given in Table 1. Figure (2) shows the top view of the eight rotors; they together make the shape of an umbrella around their local reference. Each couple of rotors is labeled by the same number and assigned to one of the main four conventional directions (1. Front_side, 3. Back_side, 4. Right_side, and 2. Left_side); on each side, the right rotor and the left rotor could also be identified. From a practical point of view, we recommend fabricating the frame of the octocopter from carbon fiber (CF) with a thickness of 4mm; this material shows high stiffness and strength with a low density that makes it at least 40% lighter than aluminum and steel; in such a case, the total weight of the basic frame of our design will be around of 4200 gr.

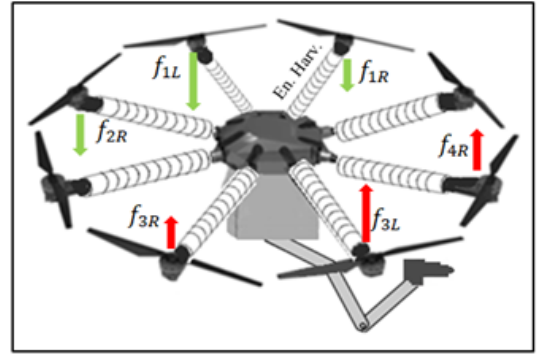

**Figure 1:** An octocopter is equipped with its reservoir. An energy harvester is installed on each rotor arm. The robot arm and the sprayer end effector are also shown. When the robot arm moves in any direction to perform a required mission, the rotors on the same side that overlap the movement of the robot arm have to increase their speed to produce sufficient lift forces to compensate for the weight which suddenly appears there. While the rotors' speed on the opposite side drops by the same corresponding values to stabilize the flying robot.

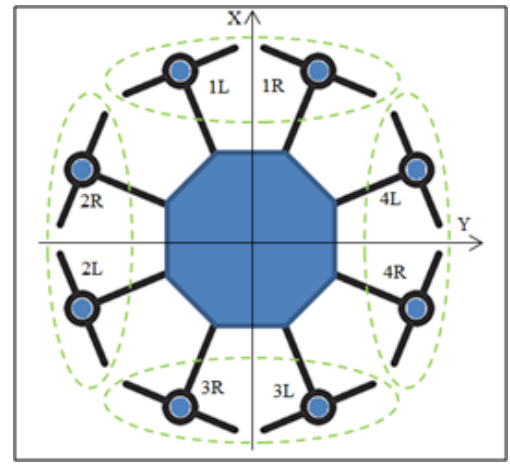

**Figure 2:** A top view of the octocopter clarifies the distribution of the eight rotors around the local reference connected to its center of gravity. Each couple of rotors is labeled by the same number and assigned to one of the main four conventional directions (1.Front_side, 3.Back_side, 4.Right_side, and 2.Left_side); on each side, we can identify the right rotor and the left rotor.

### 3.2. The electromagnetic-based embedded harvesters

A ferromagnetic core-based wrapped coil represents an effective tool to convert magnetic energy into electrical energy (Faraday's Law of Induction) Médjahdi (2019). The magnetic flux density generated around transformation cables depends on a few natural and technical factors like air humidity, the puissance of the transferred power, the electrical resistance of the harvester, and the balance of the 3-phase current. Generally, the flux density takes a minimum value of around 5μTesla for a distance of 1 m above the ground and increases when moving up near the power lines Yuan et al. (2015). In this study, we suggest a new technique that may extend the flying period of an electrical-powered UAV by providing the eight battery with a recharge current or a usable current to drive some electronic devices when it flies near some electrical power sources by converting the generated electromagnetic field into electrical power using an appropriate energy harvester.

**Table 1**
A list of the main elements regarding the considered design

| Element | Description | Gr/piece | Pieces |
|---|---|---|---|
| HD Camera | 3840x2160 Pixels, 30 frames per second | 116 | 1 |
| Rotor | 100Kv (4 axes; 40 − 58 kg); Thrust (21600 g) | 1038 | 8 |
| Drone frame | Fiber carbon-made frame | 4200 | 1 |
| Robot arm | Fiber carbon-made arm | 2200 | 1 |
| Embedded system | Xilinx Zynq-7000 FPGA | 263 | 1 |
| Rechargeable battery | Lipo (10000mAh) | 688 | 2 |
| Altimeter and force meter sensors | 50m-200 meters 24GHz Altimeter Sensor NRA24 | 100 | 8 |
| IMU + GPS | Ellipse 2 D Dual Antenna RTK INS | 180 | 1 |
| Total approximated weight | | 16751 | 23 |

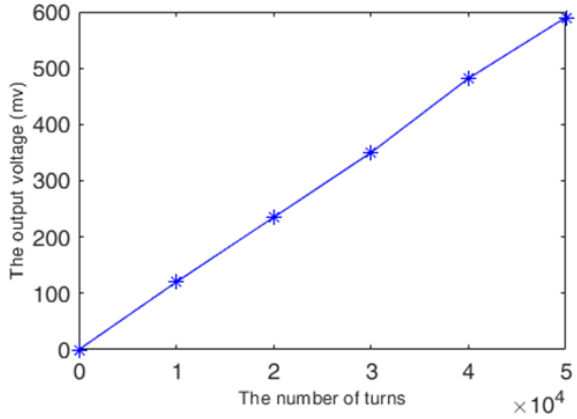

**Figure 3**: The linear relationship between the number of turns and the output voltage per harvester.

The main idea is to create a sheathe (harvester) wrapped around the eight bars that are used to install the eight rotors to the robot structure, as illustrated in figure (1). In our design, the length of each sheath is 40 centimeters and composed of 40,000 turns of a wire made of Mn-Zn (Manganese-Zinc) that has a permeability of 2300 h/m and a conductivity of 0.154 s/m. The wire's diameter is assumed of 0.14 mm, while its resistivity is 1.11 Ohm/m. The core of the harvester contains several air gaps of 0.05 mm to reduce eddy currents. According to this design, the waited output voltage from each harvester is around 481.8 mV, as reported in equation (1), while the expected power density is around 0.34mW/cm$^3$ , as given in equation (2). Figure (3) illustrates a linear relationship between the number of turns and the output voltage for each harvester.

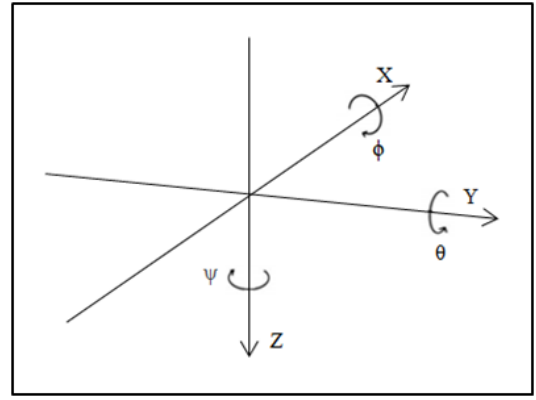

**Figure 4**: A conventional 3D local coordinate reference connected to the center of gravity of the octocopter. Three angles are considered here, Roll ($\phi$), Pitch ($\theta$), and Yaw ($\psi$).

$$V = NwAB\mu \qquad (1)$$

V is the induced voltage, N is the number of turns, w is the angular frequency (rad/s), A is the cross-section (m$^2$), B is the external magnetic flux density (Tesla), and $\mu$ is the permeability (Henry/m); the value of this last parameter is mainly related to the ferromagnetic core material as well as its shape which is the cross-section of the harvester.

$$D = \left(\frac{1}{4}\frac{V^2}{R}\right)/Har\_Vol \qquad (2)$$

D is the power density. R is the harvester coil resistance (Ohm), and Har_Vol is the harvester volume (m$^3$).

### 3.3. The Fuzzy-logic-based stabilizer

The behavior of any control system is usually described by a set of mathematical models that decide at each moment its desired output for each recently updated input, Dunn (2015). When those models become very difficult to formulate, the risk of getting a more complex system also increases. In such a case, fuzzy rules could be suggested for building a flexible linguistic terms-based system, Caponetti and Castellano (2017).

Figure (4) illustrates a conventional 3D local coordinate reference connected to the center of gravity of the octocopter, where the three Euler angles are shown; Roll ($\phi$), Pitch ($\theta$), and Yaw ($\psi$). At each moment (t), the command system of the octocopter decides three reference angles $[\theta_{ref(t)}i, \phi_{ref(t)}, \psi_{ref(t)}]$ to attain a requested attitude of the flying robot. Hence, any undesired vibration around these angles will be interpreted as an error value that is determined by calculating the difference between the reference values and the current corresponding measurements as given in the following equations for only the two considered angles Roll, and Pitch respectively:

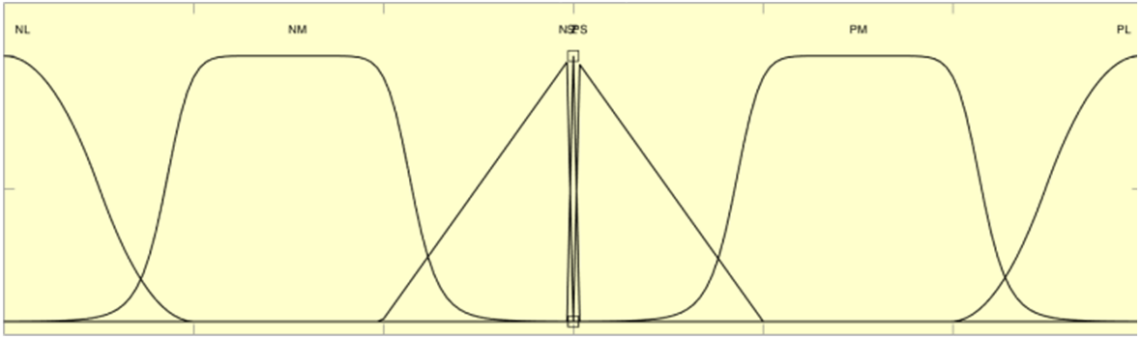

**Figure 5**: The membership functions that are used for the fuzzification and defuzzification phases. Seven fuzzy sets are considered respectively: Negative_Large (NL), Negative_Medium (NM), Negative_Small (NS), Zero (Z), Positive_Small (PS), Positive_Medium (PM), and Positive_Large (PL). The horizontal axis represents both inputs ($\Delta\theta$, $\Delta\phi$) while the vertical axis represents any of the eight outputs describing the change of the corresponding rotor speed [$\Delta$ W1R, $\Delta$ W1L, $\Delta$ W2R, $\Delta$ W2L, $\Delta$ W3R, $\Delta$ W3L, $\Delta$ W4R, $\Delta$ W4L]

$$\Delta\theta(t) = \theta_{\text{ref}}(t) - \theta_{\text{cr}}(t) \qquad (3)$$

$$\Delta\phi(t) = \phi_{\text{ref}}(t) - \phi_{\text{cr}}(t) \qquad (4)$$

$\theta_{\text{ref}}$ and $\phi_{\text{ref}}$ are two desired angles at moment (t), while $\theta_{\text{cr}}$ and $\phi_{\text{cr}}$ describe the currently measured values for both angles at the same moment.

To minimize both errors toward zero we suggest using the fuzzy logic controller as presented in our study that considers two inputs that are given in the upper-mentioned errors as in equations (3) and (4). The membership function illustrated in figure (5) plays an important role in assigning each crisp value for both inputs ($\Delta\theta$ and $\Delta\phi$) to their corresponding fuzzy values in the fuzzification phase and vice versa in the defuzzification phase. In the same figure, seven fuzzy sets could be distinguished by their corresponding crisp ranges as reported in Table 3. The reasoning phase will be performed by using the Mamdani style-based inference engine that is usually composed of fuzzy rules as given in Table 2; in each row, two fuzzy values are assigned to both inputs ($\Delta\theta$, $\Delta\phi$). Consequently, a set of predesigned values will be also assigned to each of the eight available outputs that represent the change of the corresponding rotor speed [$\Delta$W1R, $\Delta$W1L, $\Delta$W2R, $\Delta$W2L, $\Delta$W3R, $\Delta$W3L, $\Delta$W4R, $\Delta$W4L].

When the robot arm moves in an arbitrary direction to perform a required mission, the rotors that overlap on the same side ought to respond by increasing their speeds to produce sufficient lifting forces that compensate for the sudden weight appearing on the same side. While on the opposite side, the rotor's speeds have to decrease by the same corresponding values to stabilize the flying robot; figure (1). The lifting force of a rotor is usually given as follows:

$$f_{\text{i}} = k * w_i^2 \qquad (5)$$

(k) is the constant of lifting force, (w) is the rotor speed, and (i) is the serial identifier number of the rotor. As an example clarifying the functionality of this system, let's take a look at table 4, which illustrates the corresponding outputs "the changes of speed rotors in rpm" that are generated to restore the balance of the drone when two different values (in degrees) are assigned to both considered errors (inputs). In this table, we can notice that most of the positive lifting forces are provided by the rotors labeled (1L) and (2R). On the symmetric side, the compensating negative lifting forces were produced by the rotors (3L) and (4R) respectively, while the other rotors are trying to participate in small positive and negative lifting forces to get a total perfect adjustment. In this experiment, the range of rotor speed is considered between -1050 to 1050 rpm.

Compared to classical controllers we can broadly say; without looking at its linearity or even its complexity, any dynamic system can profit from the flexibility of Fuzzy logic that doesn't require a large number of parameters that should be tuned according to a set of pre-selected operating ranges of the system. The angular acceleration/deceleration of the propellers used in this design is around ±12 rad/s2. Hence, according to the experiment given in Table 4, the flying robot is able to restore its balance (getting $\Delta\theta = 0$, and $\Delta\phi = 0$) in 1.041 seconds. The same experiment could be performed by using a classical PID controller in 4.036 seconds; which means it is 4 times slower than the Fuzzy logic-based approach. All these experiments were achieved by using MATLAB – Simulink.

## 4. The computer vision-based decision-making system

As we have already reported, one of the main objectives of our design is to investigate and repair the state of the outer side of sidewalks by using a flying robot for painting the partially erased blocks that are usually supposed to be repeatedly plated in black and white. To perform this mission, the flying robot should be equipped with an embedded vision system that can detect coherent segments composed of three successive alternating patterns (white, black, white) and (black, white, black). White pixels take the intensity 1 while black pixels are supposed to be assigned to -1. Extracting any of both patterns requires applying a primitive preparation phase to the original given image. In this paper, we suggest two different experimental approaches that could be utilized to stimulate the differences among the adjacent pixels in order to lay distinct edges in between; i.e. creating all available basic segments as follows, Baba (2022):

- At first, we construct a Gabor array that extracts dissimilar available textures in the given image. Consequently, the spatially close feature sets are reshaped into an array that will be read by the PCA (Principal Component Analysis statistical procedure) to determine the main coefficients that help in creating an output image as shown in figure (7-b).

- The second method is to apply the template "MexicanHat", within the stationary wavelet transformation (SWT) technique, to the original image, where an array of coefficients for each shifted and scaled version of the original wavelet will be determined to create the output image as illustrated in figure (7-c), SWT is preferred here to keep the same resolution of the treated image.

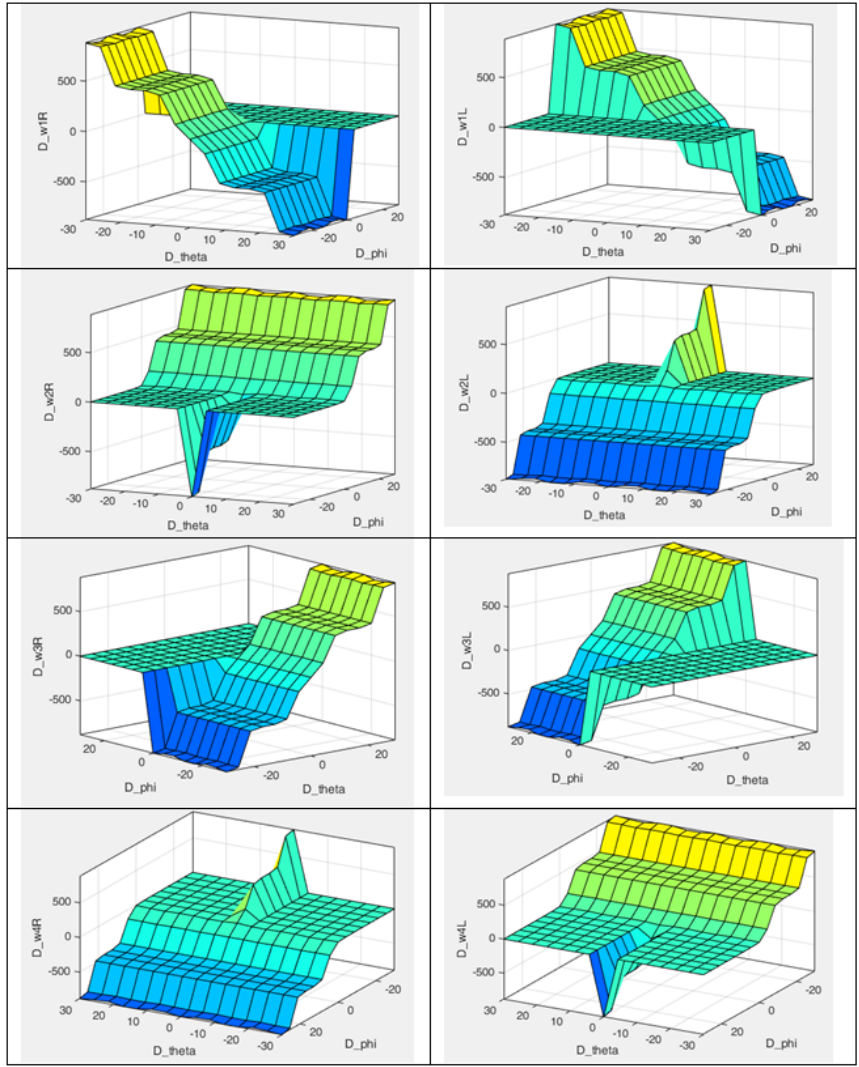

**Figure 6**: 3D surfaces showing the relationship between each considered input and each corresponding output; they interpret the available fuzzy rules.

Then, according to the vertical limits shown in the same figure (7) between each black and white segment (block), the desired patterns will be created and presented as 2D subarrays with different dimensions. To evaluate their state, i.e. to decide if they require being partially or fully plated, each pattern should be converted from an array of its original size into a vector composed of only three elements; where 1 is white, -1 is black, and 0 is any gray level in between. As shown in figure (8), this latter process considers the following heuristic concept; in one segment, if the majority of pixels are white or black, the segment will be assigned to the same intensity level (1 or -1) respectively. Otherwise, the segment's color is vague or considered partially erased, hence it will be assigned to the value 0. The data type of all pixels in the given image is scaled double. To make the rational decision, a Hopfield neural network which is composed of three neurons, as shown in figure (9), was already trained with two fundamental memories [1, -1, 1] and [-1, 1, -1]. Hence, applying one of both fundamental memories to the trained network will be directly recognized on its output at the same iteration (as they represent the correct required patterns). On the other hand, applying any partially incorrect vector like [0, -1, 1], [1, 0, 1], or [1, -1, 0] will demand more than one iteration to attract the nearest corresponding fundamental memory; in such a case and depending on the number of the required iterations to recognize the given vector that should appear on the output of the network, the flying robot becomes able to deduce a smart decision by applying the appropriate color (Black or white) to the partially erased block as shown in figure (10).

**Table 2**
The fuzzy rules of our fuzzy logic-based controller; in each row, two fuzzy values are assigned to both inputs ($\Delta\theta$, $\Delta\phi$). Consequently, a set of predesigned values will be also assigned to each of the eight available outputs that represent the change of the corresponding rotor speed [$\Delta W1R$, $\Delta W1L$, $\Delta W2R$, $\Delta W2L$, $\Delta W3R$, $\Delta W3L$, $\Delta W4R$, $\Delta W4L$]. The different colors in the table show the symmetry of the Fuzzy rules.

| Δ Theta | Δ Phi | Δ W1R | Δ W1L | Δ W2R | Δ W2L | Δ W3R | Δ W3L | Δ W4R | Δ W4L |
|---|---|---|---|---|---|---|---|---|---|
| NL | NL | PL | Z | Z | NL | NL | Z | Z | PL |
|    | NM | PL | Z | Z | NM | NL | Z | Z | PM |
|    | NS | PL | Z | Z | NS | NL | Z | Z | PS |
|    | Z  | PL | PL | Z | Z | NL | Z | Z | Z |
|    | PS | Z | PL | PS | Z | Z | NL | NS | Z |
|    | PM | Z | PL | PM | Z | Z | NL | NM | Z |
|    | PL | Z | PL | PL | Z | Z | NL | NL | Z |
| NM | NL | PM | Z | Z | NL | NM | Z | Z | PL |
|    | NM | PM | Z | Z | NM | NM | Z | Z | PM |
|    | NS | PM | Z | Z | NS | NM | Z | Z | PS |
|    | Z  | PM | PM | Z | Z | NM | NM | Z | Z |
|    | PS | Z | PM | PS | Z | Z | NM | NS | Z |
|    | PM | Z | PM | PM | Z | Z | NM | NM | Z |
|    | PL | Z | PM | PL | Z | Z | NM | NL | Z |
| NS | NL | PS | Z | Z | NL | NS | Z | Z | PL |
|    | NM | PS | Z | Z | NM | NS | Z | Z | PM |
|    | NS | PS | Z | Z | NS | NS | Z | Z | PS |
|    | Z  | PS | PS | Z | Z | NS | NS | Z | Z |
|    | PS | Z | PS | PS | Z | Z | NS | NS | Z |
|    | PM | Z | PS | PM | Z | Z | NS | NM | Z |
|    | PL | Z | PS | PL | Z | Z | NS | NL | Z |
| Z | NL | Z | Z | NL | NL | Z | Z | PL | PL |
|   | NM | Z | Z | NM | NM | Z | Z | PM | PM |
|   | NS | Z | Z | NS | NS | Z | Z | PS | PS |
|   | Z  | Z | Z | Z | Z | Z | Z | Z | Z |
|   | PS | Z | Z | PS | PS | Z | Z | NS | NS |
|   | PM | Z | Z | PM | PM | Z | Z | NM | NM |
|   | PL | Z | Z | PL | PL | Z | Z | NL | NL |
| PS | NL | NS | Z | Z | NL | PS | Z | Z | PL |
|    | NM | NS | Z | Z | NM | PS | Z | Z | PM |
|    | NS | NS | Z | Z | NS | PS | Z | Z | PS |
|    | Z  | NS | NS | Z | Z | PS | PS | Z | Z |
|    | PS | Z | NS | PS | Z | Z | PS | NS | Z |
|    | PM | Z | NS | PM | Z | Z | PS | NM | Z |
|    | PL | Z | NS | PL | Z | Z | PS | NL | Z |
| PM | NL | NM | Z | Z | NL | PM | Z | Z | PL |
|    | NM | NM | Z | Z | NM | PM | Z | Z | PM |
|    | NS | NM | Z | Z | NS | PM | Z | Z | PS |
|    | Z  | NM | NM | Z | Z | PM | PM | Z | Z |
|    | PS | Z | NM | PS | Z | Z | PM | NS | Z |
|    | PM | Z | NM | PM | Z | Z | PM | NM | Z |
|    | PL | Z | NM | PL | Z | Z | PM | NL | Z |
| PL | NL | NL | Z | Z | NL | PL | Z | Z | PL |
|    | NM | NL | Z | Z | NM | PL | Z | Z | PM |
|    | NS | NL | Z | Z | NS | PL | Z | Z | PS |
|    | Z  | NL | NL | Z | Z | PL | PL | Z | Z |
|    | PS | Z | NL | PS | Z | Z | PL | NS | Z |
|    | PM | Z | NL | PM | Z | Z | PL | NM | Z |
|    | PL | Z | NL | PL | Z | Z | PL | NL | Z |

The Hopfield network as utilized here guarantees the creation of reliable decisions in real-time. In this context, we have manually prepared a dataset composed of 106 images that have the same dimensions and resolution, all those images contain colored sidewalks of different shapes; some of them represent fully colored and correct samples while the others are partially erased. When the suggested technique was applied to the dataset, it was able to distinguish the correct patterns from the first iteration by 99% of all proposed samples, while the success ratio drops to 97% for recognizing the partially erased samples. The weight array of the Hopfield network is usually calculated as follows, Negnevitsky (2001):

$$W = \sum_{j=1}^{N} V_j V_j^T - N * I(N) \qquad (6)$$

**Table 3**
The available fuzzy sets with their corresponding crisp ranges.

| The fuzzy set | The crisp range in degrees |
|---|---|
| Negative_Large (NL) | [-30, -20] |
| Negative_Medium (NM) | [-25, -5] |
| Negative_Small (NS) | [-10, 0] |
| Zero (Z) | 0 |
| Positive_Small (PS) | [0, 10] |
| Positive_Medium (PM) | [5, 25] |
| Positive_Large (PL) | [20, 30] |

**Table 4**
As an example, the corresponding crisp outputs are calculated and given in (rpm and rad/sec) when a perturbation is considered and applied to both inputs in degrees.

| The rotor speed change | Δθ=-3.2 degrees, Δφ=1.7 degrees | |
|---|---|---|
| | rpm | rad/sec |
| ΔW1R | 2.62 | 0.274 |
| ΔW1L | 120 | 12.552 |
| ΔW2R | 116 | 12.1336 |
| ΔW2L | -5.05 | -0.528 |
| ΔW3R | -2.62 | -0.274 |
| ΔW3L | -120 | -12.552 |
| ΔW4R | -116 | -12.1336 |
| ΔW4L | 5.05 | 0.528 |

($N$) is the number of vectors to be memorized, ($V_j$ and $V_j^T$) are the considered vectors and their transposed values, respectively. ($I$) is the identity matrix with ones on the main diagonal.

Finally, all the aforementioned subsystems and some other mechanical and electronic parts should be integrated into only one embedded system that can respond in real-time to each input signal coming from the different available sensors. In this context, different platforms could be suggested

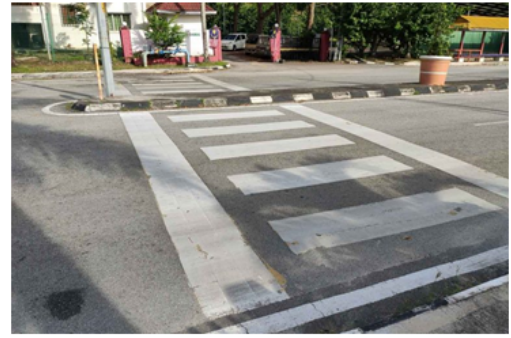

a.   The original image

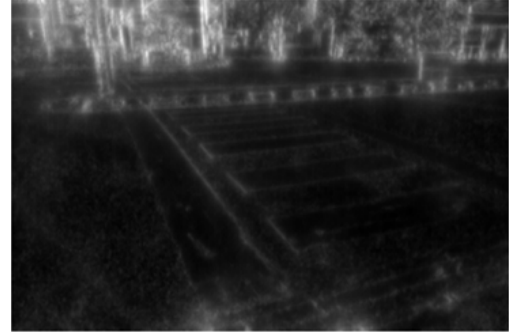

b.   Gabor filter with PCA was applied here.

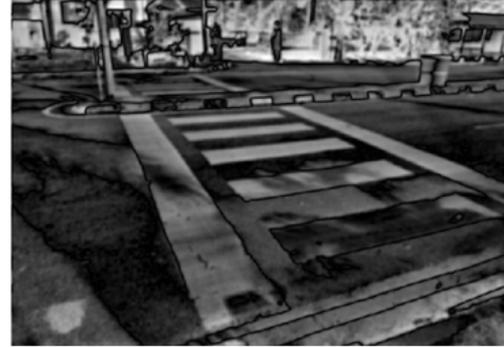

c.   "MexicanHat" template from the wavelet transformation technique was applied to the same image.

**Figure 7:** Fig.7. The original image is associated with two results illustrating how the walk side could be detected by using two different techniques: b. Gabor filter + PCA c. " MexicanHat" template; using the discrete wavelet transformation

here. To profit from its high-frequency clock signal, reliability, and reusability, we prefer employing an FPGA-based development board (Xilinx; Zynq-7000) which could be programmed by using System Verilog as a hardware description programming language.

## 5. Conclusion:

Flying robots are increasingly utilized in many areas to facilitate our daily requirements. Therefore, a lot of ideas are continuously appearing to create smart cities where vital services are supposed to be provided to people in ideal scenarios. In this context, we present a new design of a flying robot that is supposed to transport liquids and move a 3DoF robot arm in any direction while keeping perfect stability. In this study, the flying robot will be utilized to color the outer edge of a sidewalk if any of its blocks were par-tially or fully erased. As it is always regarded as much more stable compared to a quadcopter, an octocopter is utilized in our project with two main technical features. At first, we suggest a Fuzzy Logic-based stabilization system to guarantee a smooth control system that is simulated, presented in detail, and compared to PID classical controllers. Then, we improve the energy management of our flying device by proposing a full design for an embedded harvester that profits from the electromagnetic energy generated around electric transmission lines into a usable current which may increase the maximal travel distance of the robot. Finally, we briefly presented a heuristic complement study to explain a computer vision-based decision-making system used to detect the sidewalk and evaluate its coloring state (fully or partially erased), in this context several preprocessing techniques were illustrated to detect coherent segments composed of three successive alternating patterns (white, black, white) and (black, white, black). Then Hopfield neural network composed of three neurons was trained and exploited to recognize both fundamental required patterns.

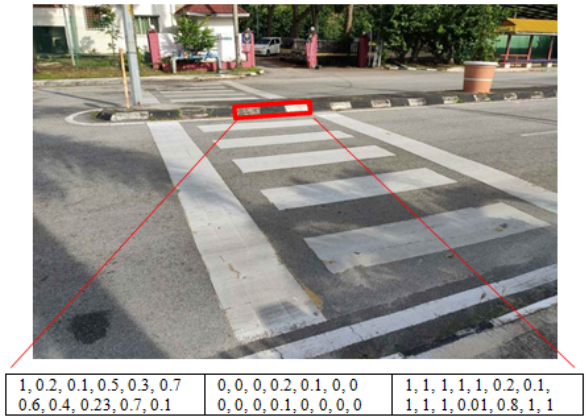

| 1, 0.2, 0.1, 0.5, 0.3, 0.7 0.6, 0.4, 0.23, 0.7, 0.1 | 0, 0, 0, 0.2, 0.1, 0, 0 0, 0, 0, 0.1, 0, 0, 0, 0 | 1, 1, 1, 1, 1, 0.2, 0.1, 1, 1, 1, 0.01, 0.8, 1, 1 |
|---|---|---|

**Figure 8:** Fig.8. For example, from right to left, white, black, and ambiguous segments are shown respectively. The three segments will be converted into the simplified vector [0, -1, 1].

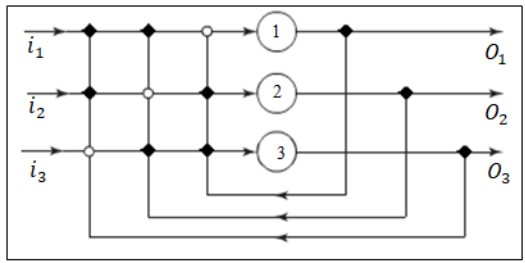

**Figure 9:** Fig.9. The Hopfield neural network is composed of three neurons; the output of each neuron is connected to the inputs of the other neurons but not connected to itself. Three inputs are ($i_1$, $i_2$, and $i_3$), while the outputs are ($O_1$, $O_2$, and $O_3$).

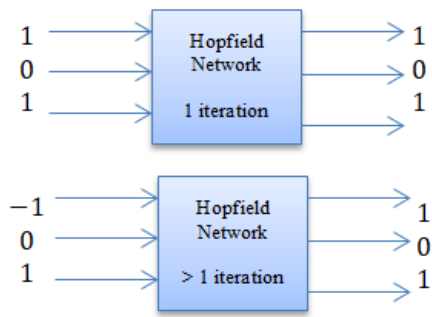

**Figure 10:** Fig.10. When the input vector is a fundamental memory, it will be recognized on the outputs of the Hopfield network in one iteration. When the input vector is partially incorrect, the nearest corresponding fundamental memory will be reorganized in more than one iteration. Hence, the system can distinguish the block to be plated.

This research did not receive any specific grant from funding agencies in the public, commercial, or not-for-profit sectors.

**Declarations of interest:** none